\documentclass[runningheads]{llncs}
\usepackage{graphicx}
\usepackage{amsmath,amssymb} 
\usepackage{color}
\usepackage{csquotes}
\usepackage{cite}
\usepackage[small]{caption}

\usepackage{subcaption}
\usepackage{booktabs}       

\usepackage{mathtools}
\usepackage{xspace}

\usepackage{hyperref}
\begin{document}
\pagestyle{headings}
\mainmatter
\def\ECCV16SubNumber{6}  

\title{Recovering 6D Object Pose: A Review and Multi-modal Analysis} 

\titlerunning{Recovering 6D Object Pose: A Review and Multi-modal Analysis}

\authorrunning{C. Sahin and T-K. Kim}

\author{Caner Sahin \and Tae-Kyun Kim}

\institute{ICVL, Imperial College London}

\maketitle

\begin{abstract}
A large number of studies analyse object detection and pose estimation at visual level in 2D, discussing the effects of challenges such as occlusion, clutter, texture, \textit{etc.}, on the performances of the methods, which work in the context of RGB modality. Interpreting the depth data, the study in this paper presents thorough multi-modal analyses. It discusses the above-mentioned challenges for full 6D object pose estimation in RGB-D images comparing the performances of several 6D detectors in order to answer the following questions: What is the current position of the computer vision community for maintaining \enquote{\textup{automation}} in robotic manipulation? What next steps should the community take for improving \enquote{\textup{autonomy}} in robotics while handling objects? Our findings include : (i) reasonably accurate results are obtained on textured-objects at varying viewpoints with cluttered backgrounds. (ii) Heavy existence of occlusion and clutter severely affects the detectors, and similar-looking distractors is the biggest challenge in recovering instances' 6D. (iii) Template-based methods and random forest-based learning algorithms underlie object detection and 6D pose estimation. Recent paradigm is to learn deep discriminative feature representations and to adopt CNNs taking RGB images as input. (iv) Depending on the availability of large-scale 6D annotated depth datasets, feature representations can be learnt on these datasets, and then the learnt representations can be customized for the 6D problem.
\end{abstract}

\section{Introduction}
\label{Introduction}
\begin{figure}[!t]
\captionsetup[subfigure]{labelformat=empty}
\centering
\includegraphics[height=4.1in]{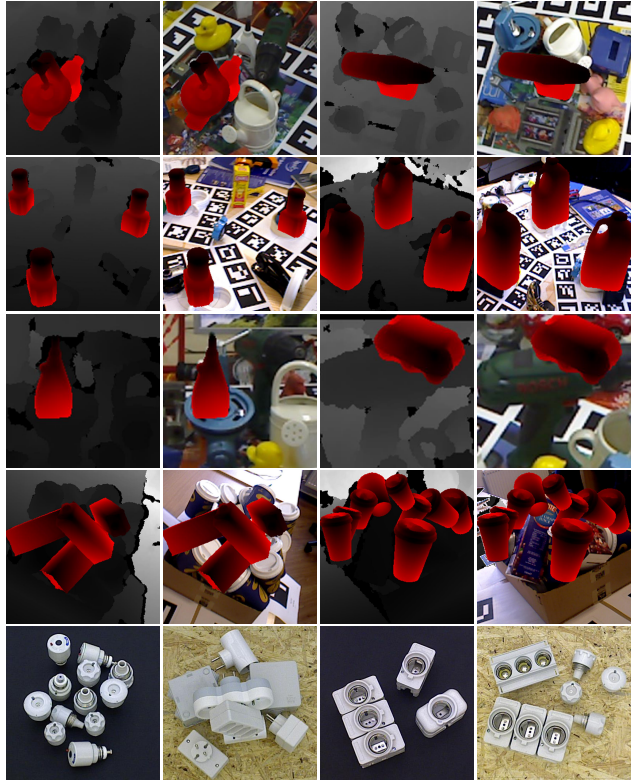}
\caption{Benchmarks collected mainly differ from the point of challenges that they involve. Row-wise, the $1^{st}$ benchmark concerns texture-less objects at varying viewpoint with cluttered background, the $2^{nd}$ is interested in multi-instance, the $3^{rd}$ has scenes with severely occluded objects, the $4^{th}$ reflects the challenges found in bin-picking scenarios, and the $5^{th}$ is related to similar-looking distractors.}
\label{fig1}
\end{figure}
Object detection and pose estimation is an important problem in the realm of computer vision, for which a large number of solutions have been proposed. One line of the solutions is based on visual perception in RGB channel. Existing evaluation studies \cite{40, 41} addressing this line of the solutions discuss the effects of challenges, such as occlusion, clutter, texture, \textit{etc}, on the performances of the methods, which are mainly evaluated on large-scale datasets, \textit{e.g.}, ImageNet \cite{45}, PASCAL \cite{46}. These studies have made important inferences for generalized object detection, however, the discussions have been restricted to visual level in 2D, since the interested methods are designed to work in the context of RGB modality.\\
\indent Increasing ubiquity of Kinect-like RGB-D sensors has prompted an interest in full 6D object pose estimation. Interpreting the depth data, state-of-the-art approaches for object detection and 6D pose estimation \cite{32, 20, 31} report improved results tackling the aforesaid challenges in 6D. This improvement is of great importance to many higher level tasks, \textit{e.g.}, scene interpretation, augmented reality, and particularly, to robotic manipulation.\\
\indent Robotic manipulators that pick and place the goods from conveyors, shelves, pallets, \textit{etc.}, can facilitate several processes comprised within logistics systems, \textit{e.g.}, warehousing, material handling, packaging. Amazon Picking Challenge (APC) \cite{16} is an important example demonstrating the promising role of robotic manipulation for the facilitation of such processes. APC integrates many tasks, such as mapping, motion planning, grasping, object manipulation, \textit{etc.}, with the goal of \enquote{\textit{autonomously}} moving items by robotic systems from a warehouse shelf into a tote \cite{14, 17}. Regarding the \enquote{\textit{automated}} handling of items by robots, accurate object detection and 6D pose estimation is an important task that when successfully performed improves the autonomy of the manipulation. Within this context, we ask the following questions. What is the current position of the computer vision community for maintaining automation in robotic manipulation, with respect to the accuracy of the 6D detectors introduced? What next steps should the community take for improving the autonomy in robotics while handling objects? We aim at answering these questions performing multi-modal analyses for object detection and 6D pose estimation where we compare state-of-the-art baselines regarding the challenges involved in the interested datasets.\\
\indent Direct comparison of the baselines is difficult, since they are tested on samples which are collected at non-identical scenarios by using RGB-D sensors with different characteristics. Additionally, different evaluation criteria are utilized for performance measure. In order to address such difficulties, we follow a threefold strategy: we firstly collect five representative object datasets \cite{32, 20, 18, 31, 61} (see Fig. \ref{fig1}). Then, we investigate $10$ state-of-the-art detectors \cite{32, 20, 31, 18, 64, 66, 21, 33, 72, 68} on the collected datasets under uniform scoring criteria of the Average Distance (AD) metric. We further extend our investigations comparing $2$ of the detectors \cite{32, 20}, which are our own implementations, using the Visible Surface Discrepancy (VSD) protocol. We offer a number of insights for the next steps to be taken, for improving the autonomy in robotics. To summarize, our main contributions are as follows:
\begin{itemize}
\item This is the first time, the current position of the field is analysed regarding object detection and 6D pose estimation.
\item We collect five representative publicly available datasets. In total, there are approximately $50$ different object classes. We investigate ten classes of the state-of-the-art 6D detectors on the collected datasets under uniform scoring criteria. 
\item We discuss baselines' strength and weakness with respect to the challenges involved in the interested RGB-D datasets. We identify the next steps for improving the robustness of the detectors, and for improving the autonomy in robotic applications, consequently.
\end{itemize}
\section{Related Work}
\label{Related_Work}
Methods producing 2D bounding box hypotheses in color images \cite{49, 50, 53, 52, 55, 51, 56, 59, 60, 54} form one line of the solutions for object detection and pose estimation. Evaluation studies interested in this line of the solutions mainly analyse the performances of the methods regarding the challenges involved within the datasets \cite{45, 46}, on which the methods have been tested. In \cite{39}, the effect of different context sources, such as geographic context, object spatial support, \textit{etc.}, on object detection is examined. Hoiem et al. \cite{40} evaluate the performances of several baselines on PASCAL dataset particularly analysing the reasons why false positives are hypothesised. Since there are less number of object categories in PASCAL dataset, Russakovsky et al. \cite{41} use ImageNet in order to do meta-analysis, and to examine the influences of color, texture, \textit{etc.}, on the performances of object detectors. Torralba et al. \cite{42} compares several datasets regarding the involved samples, cross-dataset generalization, and relative data bias, \textit{etc}. Recently published retrospective evaluation \cite{47} and benchmarking \cite{44} studies perform the most comprehensive analyses on 2D object localization and category detection, by examining the PASCAL Visual Object Classes (VOC) Challenge, and the ImageNet Large Scale Visual Recognition Challenge, respectively. These studies introduce important implications for generalized object detection, however, the discussions are restricted to visual level in 2D, since the concerned methods are engineered for color images. In this study, we target to go beyond visual perception and extend the discussions on existing challenges to 6D, interpreting depth data.\\
\begin{table*}[t]
\caption{Datasets collected: each dataset shows different characteristics mainly from the challenge point of view (VP: viewpoint, O: occlusion, C: clutter, SO: severe occlusion, SC: severe clutter, MI: multiple instance, SLD: similar looking distractors, BP: bin picking).}
\vspace{0.8em}
\centering
\setlength\tabcolsep{6pt}
\begin{minipage}{\textwidth}
\centering
\resizebox{0.99\columnwidth}{!}{
\begin{tabular}{ l c c c c c}
    \toprule
    \textbf{Dataset} & Challenge & \# Obj. Classes & Modality & \# Total Frame & Obj. Dist. [mm]\\ 
    \midrule
    LINEMOD          & VP + C + TL               &$15$ &RGB-D &15770 &600-1200 \\ 
    MULT-I           & VP + C + TL + O + MI      &$6$  &RGB-D &2067  &600-1200\\
    OCC              & VP + C + TL + SO          &$8$  &RGB-D &9209  &600-1200\\ 
    BIN-P            & VP + SC + SO + MI + BP    &$2$  &RGB-D &180   &600-1200\\
    T-LESS          & VP + C + TL + O + MI + SLD &$30$ &RGB-D &10080 &600-1200\\
    \bottomrule
  \end{tabular}
}
\label{tab_1} 
\end{minipage}%
\end{table*}
\section{Datasets}
\label{datasets}
Every dataset used in this study is composed of several object classes, for each of which a set of RGB-D test images are provided with ground truth 6D object poses. The collected datasets mainly differ from the point of the challenges that they involve (see Table \ref{tab_1}).\\
\indent \textbf{Viewpoint (VP) + Clutter (C).} Every dataset involves the test scenes in which objects of interest are located at \textit{varying viewpoints} and \textit{cluttered backgrounds}.\\
\indent \textbf{VP + C + Texture-less (TL).} Test scenes in the LINEMOD \cite{32} dataset involve \textit{texture-less} objects at varying viewpoints with cluttered backgrounds. There are 15 objects, for each of which more than $1100$ real images are recorded. The sequences provide views from $0$ - $360$ degree around the object, $0$ - $90$ degree tilt rotation, $\mp 45$ degree in-plane rotation, and $650$ mm - $1150$ mm object distance.\\
\indent \textbf{VP + C + TL + Occlusion (O) + Multiple Instance (MI).} Occlusion is one of the main challenges that makes the datasets more difficult for the task of object detection and 6D pose estimation. In addition to close and far range 2D and 3D clutter, testing sequences of the Multiple-Instance (MULT-I) dataset \cite{20} contain \textit{foreground occlusions} and \textit{multiple object instances}. In total, there are approximately $2000$ real images of $6$ different objects, which are located at the range of $600$ mm - $1200$ mm. The testing images are sampled to produce sequences that are uniformly distributed in the pose space by $[0^\circ - 360^\circ ]$, $[-80^\circ - 80^\circ ]$, and $[-70^\circ - 70^\circ ]$ in the yaw, roll, and pitch angles, respectively.\\
\indent \textbf{VP + C + TL + Severe Occlusion (SO).} Occlusion, clutter, texture-less objects, and change in viewpoint are the most well-known challenges that could successfully be dealt with the state-of-the-art 6D object detectors. However, \textit{heavy existence} of these challenges severely degrades the performance of 6D object detectors. Occlusion (OCC) dataset \cite{31} is one of the most difficult datasets in which one can observe up to  $70-80 \%$ occluded objects. OCC includes the extended ground truth annotations of LINEMOD: in each test scene of the LINEMOD \cite{32} dataset, various objects are present, but only ground truth poses for one object are given. Brachmann et al. \cite{31} form OCC considering the images of one scene (benchvise) and annotating the poses of 8 additional objects.\\
\indent \textbf{VP + SC + SO + MI + Bin Picking (BP).} In \textit{bin-picking} scenarios, multiple instances of the objects of interest are arbitrarily stocked in a bin, and hence, the objects are inherently subjected to severe occlusion and severe clutter. Bin-Picking (BIN-P) dataset \cite{18} is created to reflect such challenges found in industrial settings. It includes $183$ test images of $2$ textured objects under varying viewpoints.\\
\indent \textbf{VP + C + TL + O + MI + Similar Looking Distractors (SLD).} \textit{Similar-looking distractor(s)} along with similar looking object classes involved in the datasets strongly confuse recognition systems causing a lack of discriminative selection of shape features. Unlike the above-mentioned datasets and their corresponding challenges, the T-LESS \cite{61} dataset particularly focuses on this problem. The RGB-D images of the objects located on a table are captured at different viewpoints covering $360$ degrees rotation, and various object arrangements generate occlusion. Out-of-training objects, similar looking distractors (planar surfaces), and similar looking objects cause $6$ DoF methods to produce many false positives, particularly affecting the depth modality features. T-LESS has $30$ texture-less industry-relevant objects, and $20$ different test scenes, each of which consists of $504$ test images.\\
\section{Baselines}
\label{baselines}
State-of-the-art baselines for 6D object pose estimation address the challenges studied in Sect. \ref{datasets}, however, the architectures used differ between the baselines. In this section, we analyse 6D object pose estimators architecture-wise.\\
\indent \textbf{Template-based.} Template-based approaches, matching global descriptors of objects to the scene, are one of the most widely used approaches for object detection tasks, since they do not require time-consuming training effort. Linemod \cite{32}, being at the forefront of object detection research, estimates cluttered object's 6D pose using color gradients and surface normals. It is improved by discriminative learning in \cite{34}. Fast directional chamfer matching (FDCM) \cite{63} is used in robotics applications.\\
\indent \textbf{Point-to-point.} Point-to-point techniques build point-pair features for sparse representations of the test and the model point sets. Drost et al. \cite{64} propose create a global model description based on oriented point pair features and match that model locally using a fast voting scheme. Its further improved in \cite{66} making the method more robust across clutter and sensor noise.\\
\indent \textbf{Conventional Learning-based.} These methods are in need of training sessions where training samples along with the ground truth annotations are learnt. Latent-class Hough forests \cite{20, 73}, employing one-class learning, utilize surface normals and color gradients features in a part-based approach in order to provide robustness across occlusion. The random forest based method in \cite{31} encodes contextual information of the objects with simple depth and RGB pixels, and improves the confidence of a pose hypothesis using a Ransac-like algorithm. An analysis-by-synthesis approach \cite{36} and an uncertainty-driven methodology \cite{21} are build upon random forests, using the architecture provided in \cite{31}. The method based on random forests presented in \cite{30} formulates the recognition problem globally and derives occlusion aware features computing a set of principal curvature ratios for all pixels in depth images. The depth-based architectures in \cite{19, 67} present iterative Hough forests that initially estimate coarse 6D pose of an object, and then iteratively refine the confidence of the estimation due to the extraction of more discriminative control point descriptors \cite{71}.\\
\indent \textbf{Deep learning.} Current paradigm in the community is to learn deep discriminative feature representations. Wohlhart et al. \cite{38} utilize a CNN structure to learn discriminative descriptors and then pass the learnt descriptors to a Nearest Neighbor classifier in order to find the closest object pose. Although promising, this method has one main limitation, which is the requirement of background images during training along with the ones holistic foreground, thus making its performance dataset-specific. The studies in \cite{18, 33} learn deep representation of parts in an unsupervised fashion only from foreground images using auto-encoder architectures. The features extracted in the course of the test are fed into a Hough forest in \cite{18}, and into a codebook of pre-computed synthetic local object patches in \cite{33} in order to hypothesise object 6D pose. While \cite{38} focuses on learning feature embeddings based on metric learning with triplet comparisons, Balntas et al. \cite{74} further examine the effects of using object poses as guidance to learning robust features for 3D object pose estimation in order to handle symmetry issue.\\
\indent More recent methods adopt CNNs for 6D pose estimation, taking RGB images as inputs \cite{72}. BB8 \cite{69} and Tekin et al. \cite{70} perform corner-point regression followed by PnP for 6D pose estimation. Typically employed is a computationally expensive post processing step such as iterative closest point (ICP) or a verification network \cite{68}.\\
\section{Evaluation Metrics}
\label{Evaluation_met}
Several evaluation metrics are proposed for measuring the performance of a 6D detector. Average Distance (AD) \cite{32} outputs the score $\omega$ that calculates the distance between ground truth and estimated poses of a test object using its model. Hypotheses ensuring the following inequality is considered as correct:
\begin{equation}
\omega \leq z_{\omega} \Phi
\label{eq1}
\end{equation}
where $\Phi$ is the diameter of the 3D model of the test object, and $z_{\omega}$ is a constant that determines the coarseness of an hypothesis which is assigned as correct. Translational and rotational error function \cite{13}, being independent from the models of objects, measures the correctness of an hypothesis according to the followings: i) $\mathcal{L}_2$ norm between the ground truth and estimated translations, ii) the angle computed from the axis-angle representation of ground truth and estimated rotation matrices.\\
\indent Visible Surface Discrepancy (VSD) has recently been proposed to eliminate ambiguities arising from object symmetries and occlusions \cite{62}. The model of an object of interest is rendered at both ground truth and estimated poses, and their depth maps are intersected with the test image itself in order to compute the visibility masks. Comparing the generated masks, the score normalized in $[0 - 1]$ determines whether an estimation is correct, according to the pre-defined thresholds.\\
\indent In this study, we employ a twofold evaluation strategy for the 6D detectors using both AD and VSD metrics: i) Recall. The hypotheses on the test images of every object are ranked, and the hypothesis with the highest weight is selected as the estimated 6D pose. Recall value is calculated comparing the number of correctly estimated poses and the number of the test images of the interested object. ii) F1 scores. Unlike recall, all hypotheses are taken into account, and F1 score, the harmonic mean of precision and recall values, is presented.\\
\begin{table}[!t]
\tiny
\caption{Methods' performance are depicted object-wise based on recall values computed using the Average Distance (AD) evaluation protocol.}
\centering

\setlength\tabcolsep{2.2pt}
{\renewcommand{\arraystretch}{1.6}
\begin{subtable}{\linewidth}\centering
{\begin{tabular}[t]{|c|c| c c c c c c c c c c c c c |c|}
\hline
\textbf{Method} &ch. & ape & bvise & cam & can & cat & dril & duck & box & glue & hpunch & iron & lamp & phone & \textbf{AVER}\\  
\hline
    Kehl et al \cite{33}   &RGB-D  &96.9 &94.1  &97.7 &95.2 &97.4 &96.2 &97.3 &99.9 &78.6 &96.8 &98.7 &96.2 &92.8 &95.2\\
  LCHF \cite{20}   &RGB-D  &84	 &95    &72   &74 &91 &92 &91 &48 &55 &89 &72 &90 &69 &78.6\\
\hline
  Linemod \cite{32}   &RGB-D &95.8 &98.7  &97.5 &95.4 &99.3 &93.6 &95.9 &99.8 &91.8 &95.9 &97.5 &97.7 &93.3 &96.3\\
\hline
    Drost et al \cite{64}   &D &86.5 &70.7  &78.6 &80.2 &85.4 &87.3 &46 &97 &57.2 &77.4 &84.9 &93.3 &80.7  &78.9\\
\hline
\hline
      Kehl et al \cite{68}  &RGB   &65   &80    &78   &86 &70 &73 &66 &100  &100  &49 &78 &73 &79 &76.7\\
\hline
\end{tabular}}
\vspace{-1em}
\caption{LINEMOD dataset} \label{tab_2a}
\end{subtable}%
}

\setlength\tabcolsep{6pt}
{\renewcommand{\arraystretch}{1.6}
\begin{subtable}{\linewidth}\centering
{\begin{tabular}[t]{|c|c| c c c c c c |c|}
\hline
\textbf{Method}           &ch.   &camera  &cup  &joystick   &juice   &milk   &shampoo & \textbf{AVER}\\
\hline
 LCHF \cite{20}  &RGB-D &52.5    &99.8 &98.3       &99.3    &92.7   &97.2 &90\\
\hline
 Linemod \cite{32} &RGB-D &18.3    &99.2 &85         &51.6    &72.2   &53.1 &63.2\\
\hline
\end{tabular}}
\vspace{-1em}
\caption{MULT-I dataset}\label{tab_2b}
\end{subtable}%
}

\setlength\tabcolsep{4.8pt}
{\renewcommand{\arraystretch}{1.6}
\begin{subtable}{\linewidth}\centering
{\begin{tabular}[t]{|c|c| c c c c c c c c |c|}
\hline
\textbf{Method} &ch. &ape &can &cat &dril &duck &box &glue &hpunch &\textbf{AVER}\\ 
\hline
    Xiang et al. \cite{72}   &RGB-D &76.2	 &87.4  &52.2 &90.3 &77.7 &72.2 &76.7 &91.4 &78\\
\hline
    LCHF \cite{20}  &RGB-D  &48.0	 &79.0  &38.0 &83.0 &64.0 &11.0   &32.0 &69.0  &53\\
\hline
    Hinters et al. \cite{66} &RGB-D  &81.4	 &94.7  &55.2 &86.0 &79.7 &65.5 &52.1 &95.5 &76.3\\
\hline
    Linemod \cite{32} &RGB-D  &21.0	 &31.0  &14.0 &37.0 &42.0 &21.0 &5.0    &35.0 &25.8\\
\hline
\hline
    Xiang et al. \cite{72}   &RGB &9.6	 &45.2  &0.93 &41.4 &19.6 &22.0   &38.5 &22.1 &25\\
    \hline
\end{tabular}}
\vspace{-1em}
\caption{OCC dataset}\label{tab_2c}
\end{subtable}%
}

\setlength\tabcolsep{19.4pt}
{\renewcommand{\arraystretch}{1.6}
\begin{subtable}{\linewidth}\centering
{\begin{tabular}[t]{|c|c| c c |c|}
\hline
\textbf{Method}                &ch.      &cup         &juice  &\textbf{AVER}\\
\hline
	 LCHF \cite{20}   &RGB-D    &90.0	      &89.0   &90\\
	  Brach et al. \cite{31}   &RGB-D    &89.4        &87.6   &89\\
\hline
	Linemod \cite{32}   &RGB-D    &88.0	      &40.0   &64\\
\hline
\end{tabular}}
\vspace{-1em}
\caption{BIN-P dataset}\label{tab_2d}
\end{subtable}%
}
\label{tab_2}
\end{table}
\section{Multi-modal Analyses}
\label{Exp_Res}
We analyse ten baselines on the datasets with respect to both challenges and the architectures. Two of the baselines \cite{32, 20} are our own implementations. The color gradients and surface normal features, presented in \cite{32}, are computed using the built-in functions and classes provided by OpenCV. The features in Latent-Class Hough Forest (LCHF) \cite{20} are the part-based version of the features introduced in \cite{32}. Hence, we inherit the classes given by OpenCV in order to generate part-based features used in LCHF. We train each method for the objects of interest by ourselves, and using the learnt classifiers, we test those on all datasets. Note that, the methods use only foreground samples during training/template generation. In this section, \enquote{LINEMOD} refers to the dataset, whilst \enquote{Linemod} is used to indicate the baseline itself.
\subsection{Analyses Based on Average Distance}
Utilizing the AD metric, we compare the chosen baselines along with the challenges, i) regarding the recall values that each baseline generates on every dataset, ii) regarding the F1 scores. The coefficient $z_{\omega}$ is $0.10$, and in case we use different thresholds, we will specifically indicate in the related parts.
\begin{figure}[!t]
\centering
\includegraphics[height=3.2in]{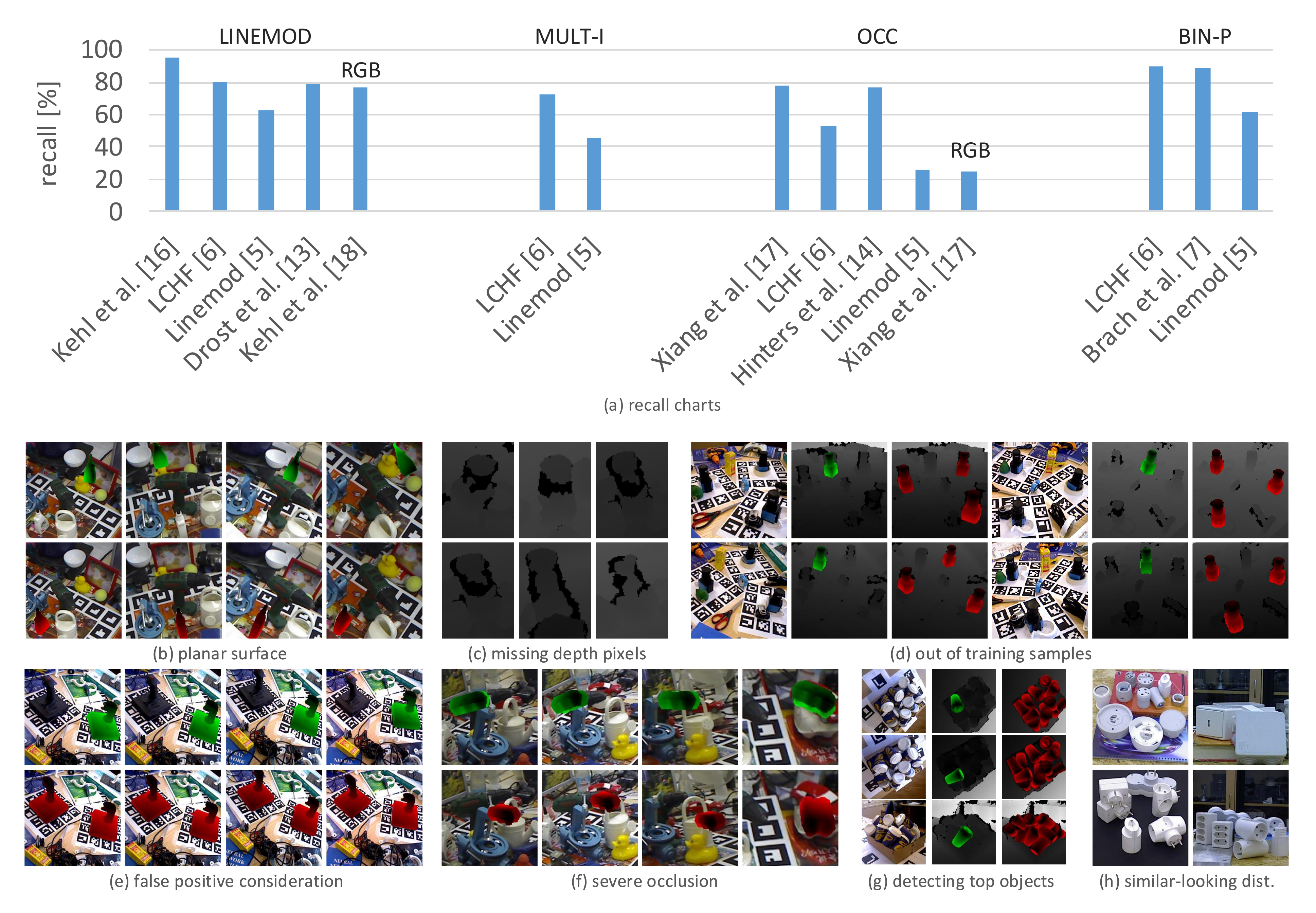}
\caption{(a) Success of each baseline on every dataset is shown, recall values are computed using the Average Distance (AD) metric. (b)-(h) challenges encountered during test are exemplified (green renderings are hypotheses, and the red ones are ground truths).}
\label{fig2}
\end{figure}
\subsubsection{Recall-only Discussions}
\label{sub_recal}
Recall-only discussions are based on the numbers provided in Table \ref{tab_2}, and Fig. \ref{fig2}.\\
\indent \textbf{Clutter, Viewpoint, Texture-less objects.} Highest recall values are obtained on the LINEMOD dataset (see Fig. \ref{fig2} (a)), meaning that the state-of-the-art methods for 6D object pose estimation can successfully handle the challenges, clutter, varying viewpoint, and texture-less objects. LCHF, detecting more than half of the objects with over $80 \%$ accuracy, worst performs on \enquote{box} and \enquote{glue} (see Table \ref{tab_2a}), since these objects have planar surfaces, which confuses the features extracted in depth channel (example images are given in Fig. \ref{fig2} (b)).\\
\indent \textbf{Occlusion.} In addition to the challenges involved in LINEMOD, occlusion is introduced in MULT-I. Linemod's performance decreases, since occlusion affects holistic feature representations in color and depth channels. LCHF performs better on this dataset than Linemod. Since LCHF is trained using the parts coming from positive training images, it can easily handle occlusion, using the information acquired from occlusion-free parts of the target objects. However, LCHF degrades on \enquote{camera}. In comparison with the other objects in the dataset, \enquote{camera} has relatively smaller dimensions. In most of the test images, there are non-negligible amount of missing depth pixels (Fig. \ref{fig2} (c)) along the borders of this object, and thus confusing the features extracted in depth channel. In such cases, LCHF is liable to detect similar-looking out of training objects and generate many false positives (see Fig. \ref{fig2} (d)). The hypotheses produced by LCHF for \enquote{joystick} are all considered as false positive (Fig. \ref{fig2} (e)). When we re-evaluate the recall that LCHF produces on the \enquote{joystick} object setting $z_{\omega}$ to the value of $0.15$, we observe $89 \%$ accuracy.\\
\indent \textbf{Severe Occlusion.} OCC involves challenging test images where the objects of interest are cluttered and severely occluded. The best performance on this dataset is caught by Xiang et al. \cite{72}, and there is still room for improvement in order to fully handle this challenge. Despite the fact that the distinctive feature of this benchmark is the existence of \enquote{severe occlusion}, there are occlusion-free target objects in several test images. In case the test images of a target object include unoccluded and/or naively occluded samples (with the occlusion ratio up to $40 \% -50 \%$ of the object dimensions) in addition to severely occluded samples, methods produce relatively higher recall values (\textit{e.g.} \enquote{can, driller, duck, holepuncher}, Table \ref{tab_2c}). On the other hand, when the target object has additionally other challenges such as planar surfaces, methods' performance (LCHF and Linemod) decreases (\textit{e.g.} \enquote{box}, Fig. \ref{fig2} (f)).\\
\indent \textbf{Severe Clutter.} In addition to the challenges discussed above, BIN-P inherently involves severe clutter, since it is designed for bin-picking scenarios, where objects are arbitrarily stacked in a pile. According to the recall values presented in Table \ref{tab_2d}, LCHF and Brachmann et al. \cite{31} perform $ 25 \%$ better than Linemod. Despite having severely occluded target objects in this dataset, there are unoccluded/relatively less occluded objects at the top of the bin. Since our current analyses are based on the top hypothesis of each method, the produced success rates show that the methods can recognize the objects located on top of the bin with reasonable accuracy (Fig. \ref{fig2} (g)).\\
\indent \textbf{Similar-Looking Distractors.} We test both Linemod and LCHF on the T-LESS dataset. Since most of the time the algorithms fail, we do not report quantitative analyses, instead we discuss our observations from the experiments. The dataset involves various object classes with strong shape and color similarities. When the background color is different than that of the objects of interest, color gradient features are successfully extracted. However, the scenes involve multiple instances, multiple objects similar in shape and color, and hence, the features queried exist in the scene at multiple locations. The features extracted in depth channel are also severely affected from the lack of discriminative selection of shape information. When the objects of interest have planar surfaces, the detectors cannot easily discriminate foreground and background in depth channel, since these objects in the dataset are relatively smaller in dimension (see Fig. \ref{fig2} (h)).\\
\indent \textbf{Part-based vs. Holistic approaches.} Holistic methods \cite{32, 64, 66, 72, 68} formulate the detection problem globally. Linemod \cite{32} represents the windows extracted from RGB and depth images by the surface normals and color gradients features. Distortions along the object borders arising from occlusion and clutter, that is, the distortions of the color gradient and surface normal information in the test processes, mainly degrade the performance of this detector. Part-based methods \cite{20, 31, 18, 21, 33} extract parts in the given image. Despite the fact that LCHF uses the same kinds of features as in Linemod, LCHF detects objects extracting parts, thus making the method more robust to occlusion and clutter.\\
\indent \textbf{Template-based vs. Random forest-based.} Template-based methods, \textit{i.e.}, Linemod, match the features extracted during test to a set of templates, and hence, they cannot easily be generalized well to unseen ground truth annotations, that is, the translation and rotation parameters in object pose estimation. Methods based on random forests \cite{20, 31, 18, 21} efficiently benefit the randomisation embedded in this learning tool, consequently providing good generalisation performance on new unseen samples.\\
\indent \textbf{RGB-D vs. Depth.} Methods utilizing both RGB and depth channels demonstrate higher recall values than methods that are of using only depth, since RGB provides extra clues to ease the detection. This is depicted in Table \ref{tab_2a} where learning- and template-based methods of RGB-D perform much better than point-to-point technique \cite{64} of depth channel.\\
\indent \textbf{RGB-D vs. RGB (CNN structures).} More recent paradigm is to adopt CNNs to solve 6D object pose estimation problem taking RGB images as inputs \cite{72, 68}. Methods working in the RGB channel in Table \ref{tab_2} are based on CNN structure. According to the numbers presented in Table \ref{tab_2}, RGB-based SSD-6D \cite{69} and RGB-D-based LCHF achieve similar performance. These recall values show the promising performance of CNN architectures across random forest-based learning methods.\\
\begin{table}[!t]
\tiny
\caption{Methods' performance are depicted object-wise based on F1 scores computed using the Average Distance (AD) evaluation protocol.}
\centering

\setlength\tabcolsep{2.2pt}
{\renewcommand{\arraystretch}{1.6}
\begin{subtable}{\linewidth}\centering
{\begin{tabular}[t]{|c|c| c c c c c c c c c c c c c |c|}
\hline
\textbf{Method} &ch. & ape & bvise & cam & can & cat & dril & duck & box & glue & hpunch & iron & lamp & phone & \textbf{AVER}\\ 
\hline
    Kehl et al. \cite{33}  &RGB-D &0.98 &0.95 &0.93 &0.83 &0.98 &0.97 &0.98 &1 &0.74 &0.98 &0.91 &0.98 &0.85 &0.93\\
    LCHF \cite{20}  &RGB-D &0.86 &0.96 &0.72 &0.71 &0.89 &0.91 &0.91 &0.74 &0.68 &0.88 &0.74 &0.92 &0.73 &0.82\\
\hline
    Linemod \cite{32} &RGB-D  &0.53 &0.85 &0.64 &0.51 &0.66 &0.69 &0.58 &0.86 &0.44 &0.52 &0.68 &0.68 &0.56 &0.63\\
\hline
\hline
    Kehl et al. \cite{68}  &RGB &0.76 &0.97 &0.92 &0.93 &0.89 &0.97 &0.80 &0.94 &0.76 &0.72 &0.98 &0.93 &0.92 &0.88\\
\hline
\end{tabular}}
\caption{LINEMOD dataset}\label{tab_3a}
\end{subtable}%
}

\setlength\tabcolsep{5.8pt}
{\renewcommand{\arraystretch}{1.6}
\begin{subtable}{\linewidth}\centering
{\begin{tabular}[t]{|c|c| c c c c c c |c|}
\hline
\textbf{Method}      &ch.         &camera  &cup  &joystick  &juice  &milk  &shampoo & \textbf{AVER}\\
\hline
Kehl et al. \cite{33}     &RGB-D     &0.38  &0.97  &0.89	 &0.87  &0.46  &0.91 &0.75\\
LCHF \cite{20}   &RGB-D     &0.39  &0.89  &0.55	 &0.88  &0.40  &0.79 &0.65\\
\hline
Drost et al. \cite{64}    &D     &0.41  &0.87  &0.28	 &0.60  &0.26  &0.65 &0.51\\
\hline
Linemod \cite{32}  &RGB-D     &0.37  &0.58  &0.15     &0.44  &0.49  &0.55 &0.43\\	
\hline
\hline
Kehl et al. \cite{68}     &RGB     &0.74  &0.98  &0.99  &0.92  &0.78  &0.89 &0.88\\
\hline
\end{tabular}}
\vspace{-1em}
\caption{MULT-I dataset}\label{tab_3b}
\end{subtable}%
}

\setlength\tabcolsep{4.8pt}
{\renewcommand{\arraystretch}{1.6}
\begin{subtable}{\linewidth}\centering
{\begin{tabular}[t]{|c|c| c c c c c c c c |c|}
\hline
\textbf{Method} &ch. &ape &can &cat &dril &duck &box &glue &hpunch & \textbf{AVER}\\ 
\hline
   LCHF \cite{20}  &RGB-D  &0.51	 &0.77  &0.44 &0.82 &0.66 &0.13 &0.25 &0.64 &0.53\\
\hline
   Linemod \cite{32}  &RGB-D  &0.23	 &0.31  &0.17 &0.37 &0.43 &0.19 &0.05 &0.30 &0.26\\
\hline
\hline
  Brach et al. \cite{21}   &RGB  &- &- &- &- &- &- &- &- &0.51\\
\hline
   Kehl et al. \cite{68}   &RGB  &- &- &- &- &- &- &- &- &0.38\\
\hline
\end{tabular}}
\vspace{-1em}
\caption{OCC dataset}\label{tab_3c}
\end{subtable}%
}

\setlength\tabcolsep{16.7pt}
{\renewcommand{\arraystretch}{1.6}
\begin{subtable}{\linewidth}\centering
{\begin{tabular}[t]{|c|c| c c |c|}
\hline
\textbf{Method}              &ch.    &cup   &juice &\textbf{AVER}\\
\hline
           LCHF \cite{20}    &RGB-D  &0.48  &0.29  &0.39\\
    Doumanoglou et al. \cite{18}    &RGB-D  &0.36  &0.29  &0.33\\
\hline   
        Linemod \cite{32}    &RGB-D  &0.48  &0.20  &0.34\\
\hline
\end{tabular}}
\vspace{-1em}
\caption{BIN-P dataset}\label{tab_3d}
\end{subtable}%
}
\label{tab_3}
\end{table}
\indent Robotic manipulators that pick and place the items from conveyors, shelves, pallets, \textit{etc.}, need to know the pose of one item per RGB-D image, even though there might be multiple items in its workspace. Hence our recall-only analyses mainly target to solve the problems that could be encountered in such cases. Based upon the analyses currently made, one can make important implications, particularly from the point of the performances of the detectors. On the other hand, recall-based analyses are not enough to illustrate which dataset is more challenging than the others. This is especially true in crowded scenarios where multiple instances of target objects are severely occluded and cluttered. Therefore, in the next part, we discuss the performances of the baselines from another aspect, regarding precision-recall curves and F1 scores, where the 6D detectors are investigated sorting all detection scores across all images.
\begin{figure*}[!t]
\captionsetup[subfigure]{labelformat=empty}
\centering
\includegraphics[height=3.2in]{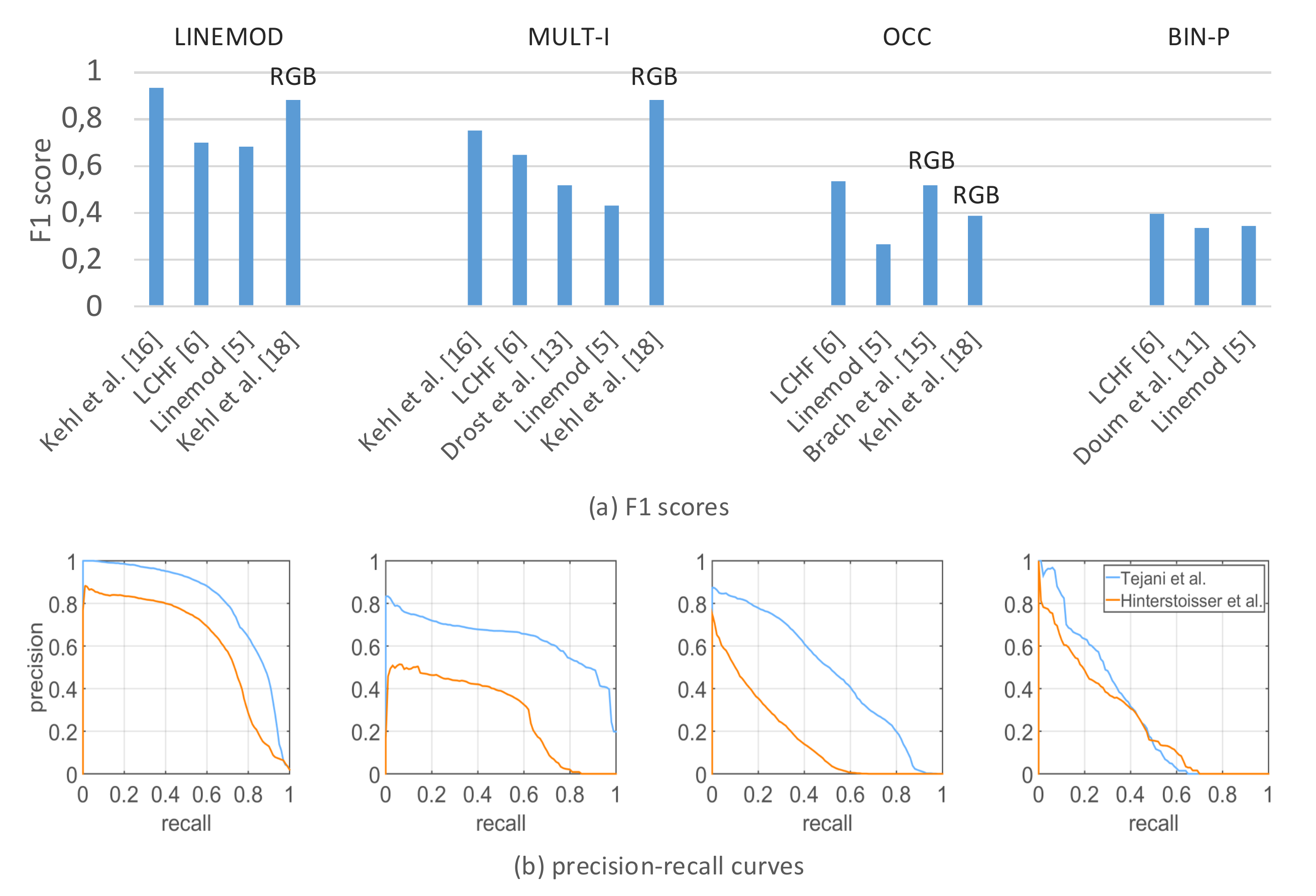}
\caption{(a) Success of each baseline on every dataset is shown, F1 scores are computed using the Average Distance (AD) metric. (b) Precision-recall curves of averaged F1 scores for Tejani et al. \cite{20} and Hinterstoisser et al. \cite{32} are shown: from left to right, LINEMOD, MULT-I, OCC, BIN-P.}
\label{fig3}
\end{figure*}
\subsubsection{Precision-Recall Discussions}
\label{sub_pr}
Our precision-recall discussions are based on the F1 scores provided in Table \ref{tab_3}, and Fig. \ref{fig3} (a).\\
\indent We first analyse the performance of the methods \cite{32, 20, 68, 65} on the LINEMOD dataset. On the average, Kehl et al. \cite{65} outperforms other methods proving the superiority of learning deep features. Despite estimating 6D in RGB images, SSD-6D \cite{68} exhibits the advantages of using CNN structures for 6D object pose estimation. LCHF and Linemod demonstrate lower performance, since the features used by these methods are manually-crafted. The comparison between Fig. \ref{fig2} (a) and Fig. \ref{fig3} (a) reveals that the results produced by the methods have
approximately the same characteristics on the LINEMOD dataset, with respect to recall and F1 scores.\\
\indent The methods tested on the MULT-I dataset \cite{65, 32, 20, 64} utilize the geometry information inherently provided by depth images. Despite this fact, SSD-6D \cite{68}, estimating 6D pose only from RGB images, outperforms other methods clearly proving the superiority of using CNNs for the 6D problem over other structures.\\
\indent LCHF \cite{32} and Brachmann et al. \cite{21} best perform on OCC with respect to F1 scores. As this dataset involves test images where highly occluded objects are located, the reported results depict the importance of designing part-based solutions.\\
\indent The most important difference is observed on the BIN-P dataset. While the success rates of the detectors on this dataset are higher than $60 \%$ with respect to the recall values (see Fig. \ref{fig2} (a)), according to the presented F1 scores, their performance are less than $40 \%$. When we take into account all hypotheses and the challenges particular to this dataset, which are severe occlusion and severe clutter, we observe strong degradation in the accuracy of the detectors.\\
\indent In Fig. \ref{fig3} (b), we lastly report precision-recall curves of LCHF and Linemod. Regarding these curves, one can observe that as the datasets are getting more difficult, from the point of challenges involved, the methods produce less accurate results.\\
\begin{figure}[!t]
\captionsetup[subfigure]{labelformat=empty}
\centering
\includegraphics[height=2.1in]{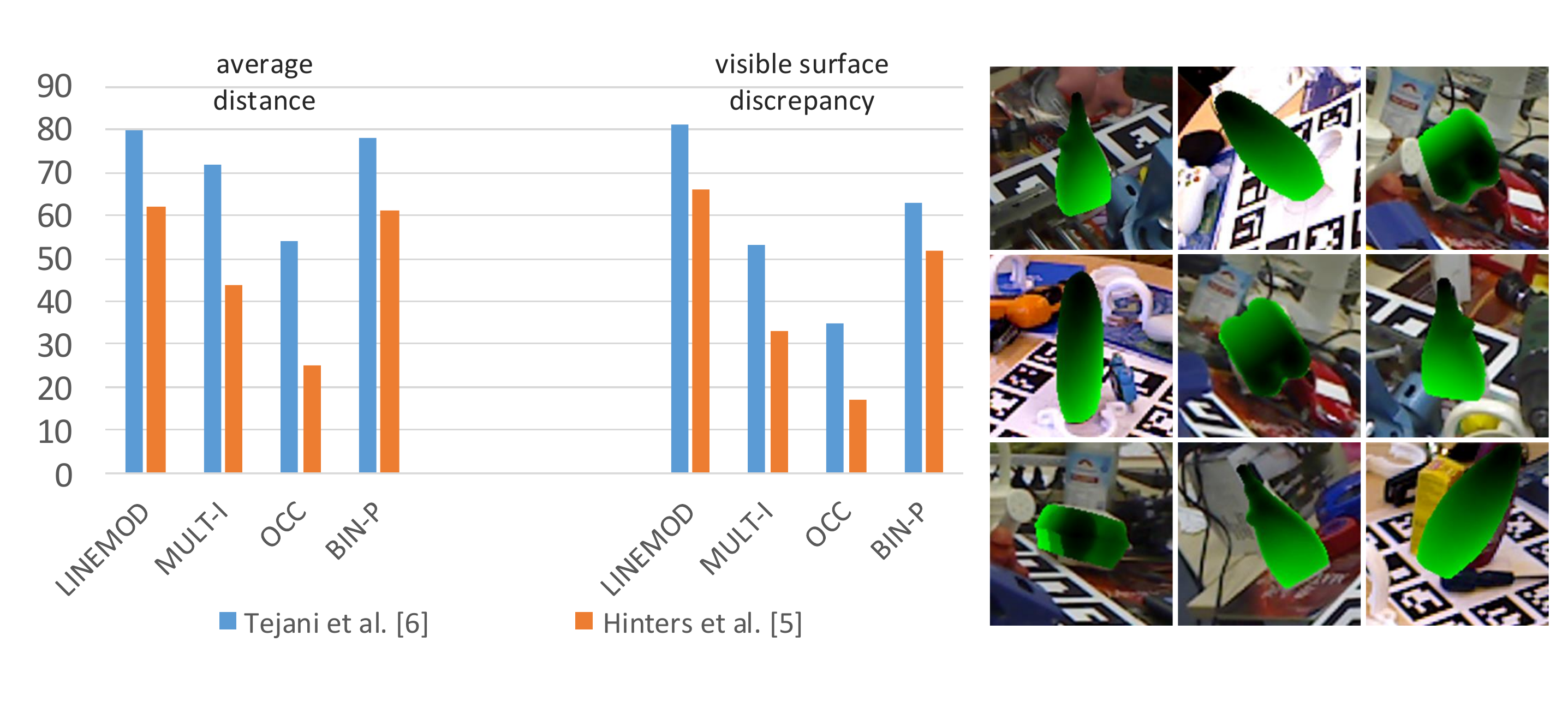}
\vspace{-1em}
\caption{Methods are evaluated based on Visible Surface Discrepancy. Samples on the right are considered as false positive with respect to Average Distance, whilst VSD deems correct.}
\vspace{-2em}
\label{fig5}
\end{figure}
\subsection{Analyses Based on Visible Surface Discrepancy}
The analyses presented so far have been employed using the AD metric. We continue our discussions computing the recall values using the VSD metric, which is inherently proposed for tackling the pose-ambiguities arising from symmetry. We set $\delta$, $\tau$, and $t$, the thresholds defined in \cite{62}, to the values of $20$ mm, $100$ mm, and $0.5$ respectively. Figure \ref{fig5} shows the accuracy of each baseline on the LINEMOD, MULT-I, OCC, BIN-P datasets, respectively. Comparing the numbers in this chart, one can observe that the results from VSD are relatively lower than that are of the AD metric. This arises mainly from the chosen parameters. However, the characteristics of both charts are the same, that is, both methods, according to AD and VSD, perform best on the LINEMOD dataset, whilst worst on OCC. On the other hand, the main advantage of the proposed metric is that it features ambiguity-invariance: Since it is designed to evaluate the baselines over the visible parts of the objects, it gives more robust measurements across symmetric objects. Sample images in Fig. \ref{fig5} show the hypotheses of symmetric objects which are considered as false positive according to the AD metric, whilst VSD accepts those as correct.
\section{Discussions and Conclusions}
We outline our key observations that provide guidance for future research.\\
\indent From the challenges aspect, reasonably accurate results have been obtained on textured-objects at varying viewpoints with cluttered backgrounds. In case occlusion is introduced in the test scenes, depending on the architecture of the baseline, good performance demonstrated. Part-based solutions can handle the occlusion problem better than the ones global, using the information acquired from occlusion-free parts of the target objects. However, heavy existence of occlusion and clutter severely affects the detectors. It is possible that modelling occlusion during training can improve the performance of a detector across severe occlusion. But when occlusion is modelled, the baseline could be data-dependent. In order to maintain the generalization capability of the baseline contextual information can additionally be utilized during the modelling. Currently, similar looking distractors along with similar looking object classes seem the biggest challenge in recovering instances' 6D, since the lack of discriminative selection of shape features strongly confuse recognition systems. One possible solution could be considering the instances that have strong similarity in shape in a same category. In such a case, detectors trained using the data coming from the instances involved in the same category can report better detection results.\\
\indent Architecture-wise, template-based methods, matching model features to the scene, and random forest based learning algorithms, along with their good generalization performance across unseen samples, underlie object detection and 6D pose estimation. Recent paradigm in the community is to learn deep discriminative feature representations. Despite the fact that several methods addressed 6D pose estimation utilizing deep features \cite{18, 33}, end-to-end neural network-based solutions for 6D object pose recovery are still not widespread. Depending on the availability of large-scale 6D annotated depth datasets, feature representations can be learnt on these datasets, and then the learnt representations can be customized for the 6D problem.\\
\indent These implications are related to automation in robotic systems. The implications can provide guidance for robotic manipulators that pick and place the items from conveyors, shelves, pallets, \textit{etc}. Accurately detecting objects and estimating their fine pose under uncontrolled conditions improves the grasping capability of the manipulators. Beyond accuracy, the baselines are expected to show real-time performance. Although the detectors we have tested cannot perform real-time, their run-time can be improved by utilizing APIs like OpenMP.

\end{document}